\documentclass{CUP-JNL-EDS}%

\usepackage{latexsym}
\usepackage{graphicx}
\usepackage{multicol,multirow}
\usepackage{amsmath,amssymb,amsfonts}
\usepackage{mathrsfs}
\usepackage{amsthm}
\usepackage{rotating}
\usepackage{appendix}
\usepackage{ifpdf}
\usepackage[T1]{fontenc}
\usepackage{times}
\usepackage{newtxmath}
\usepackage{textcomp}%
\usepackage{xcolor}%
\usepackage{hyperref}
\usepackage{lipsum}
\usepackage[noabbrev,capitalise]{cleveref}
\usepackage{dsfont}
\usepackage[]{mdframed} %
\usepackage{algorithm}
\usepackage{algpseudocode}
\usepackage{wrapfig}
\usepackage{placeins}
\usepackage{booktabs}
\usepackage{tikz}

\usepackage{bm}

\usepackage[
    style=authoryear,
    maxcitenames=2,
    mincitenames=1,
    maxbibnames=99,
    uniquelist=minyear,
    uniquename=init,
    giveninits,
    dashed=false,
]{biblatex}
\DeclareNameAlias{sortname}{family-given}
\AtEveryBibitem{\clearfield{month}}
\AtEveryBibitem{\clearfield{day}}
\AtEveryBibitem{\ifentrytype{unpublished}{}{\clearfield{eprint}\clearfield{eprinttype}}}
\DeclareSourcemap{
	\maps[datatype=bibtex, overwrite]{
		\map{
			\step[fieldset=doi, null]
			\step[fieldset=issn, null]
			\step[fieldset=isbn, null]
			\step[fieldset=editor, null]
			\step[fieldset=publisher, null]
			\step[fieldset=location, null]
			\step[fieldset=organization, null]
			\step[fieldset=series, null]
		}
	}
}
\usepackage{xcolor}

\renewcommand\vec[1]{\boldsymbol{#1}}
\newcommand\mat[1]{\boldsymbol{#1}}

\newcommand{\Reals}{\mathbb{R}}
\DeclareMathOperator*{\argmax}{arg\,max}

\newcommand{\OLIf}[2]{\State \textbf{if} #1 \textbf{then} #2}

\newtheorem{theorem}{Theorem}[section]

\theoremstyle{definition}
\newtheorem{definition}[theorem]{Definition}

\newtheorem{example}[theorem]{Example}
\theoremstyle{remark}

\articletype{METHODS PAPER}
\jname{Environmental Data Science}
\jyear{2024}
\jvol{4}
\jdoi{10.1017/eds.2020.xx}

\begin{document}

\begin{Frontmatter}

\title{Scalable Data Assimilation with Message Passing}

\author*[1]{Oscar Key*}\email{oscar.key.20@ucl.ac.uk}\orcid{0009-0009-1357-471X}
\author[1,2]{So Takao*}
\author[1]{Daniel Giles*}\orcid{0000-0002-3668-1851}
\author[1,3]{Marc Peter Deisenroth}\orcid{0000-0003-1503-680X}

\address*[*]{Equal contribution}
\address[1]{\orgdiv{UCL Centre for Artificial Intelligence}, \orgname{University College London}, \country{United Kingdom}}
\address*[2]{\orgdiv{Department of Computing and Mathematical Sciences}, \orgname{California Institute of Technology},
\state{CA},  \country{United States}}
\address*[3]{\orgname{The Alan Turing Institute}, \orgaddress{ \country{United Kingdom}}}

\received{9th February 2024}
\revised{26 April 2024}
\accepted{21 June 2024}

\authormark{Key O.*, Takao S.*, Giles D.*, Deisenroth M. P.}

\keywords{data assimilation, Bayesian inference, message passing, distributed computation}

\abstract{
    Data assimilation is a core component of numerical weather prediction systems.
    The large quantity of data processed during assimilation requires the computation to be distributed across increasingly many compute nodes, yet existing approaches suffer from synchronisation overhead in this setting.
    In this paper, we exploit the formulation of data assimilation as a Bayesian inference problem and apply a message-passing algorithm
    to solve the spatial inference problem. %
    Since message passing is inherently based on local computations, this approach lends itself to parallel and distributed computation.
    In combination with a GPU-accelerated implementation, we can scale the algorithm to very large grid sizes while retaining good accuracy and compute and memory requirements.
}

\policy{
    This paper addresses scalability issues with one of the core algorithms in numerical weather prediction systems.
    Solving these issues contributes to producing higher resolution and more frequently updated weather forecasts.
    Improved forecasts are an important tool for mitigating and adapting to climate change, with applications, such as predicting the output of wind and solar power, and warning about extreme weather events.
}

\end{Frontmatter}

\section{Introduction}
\begin{wrapfigure}{0.25\textwidth}{7cm}
    \vspace{-4ex}
    \centering
    \includegraphics[width=\hsize]{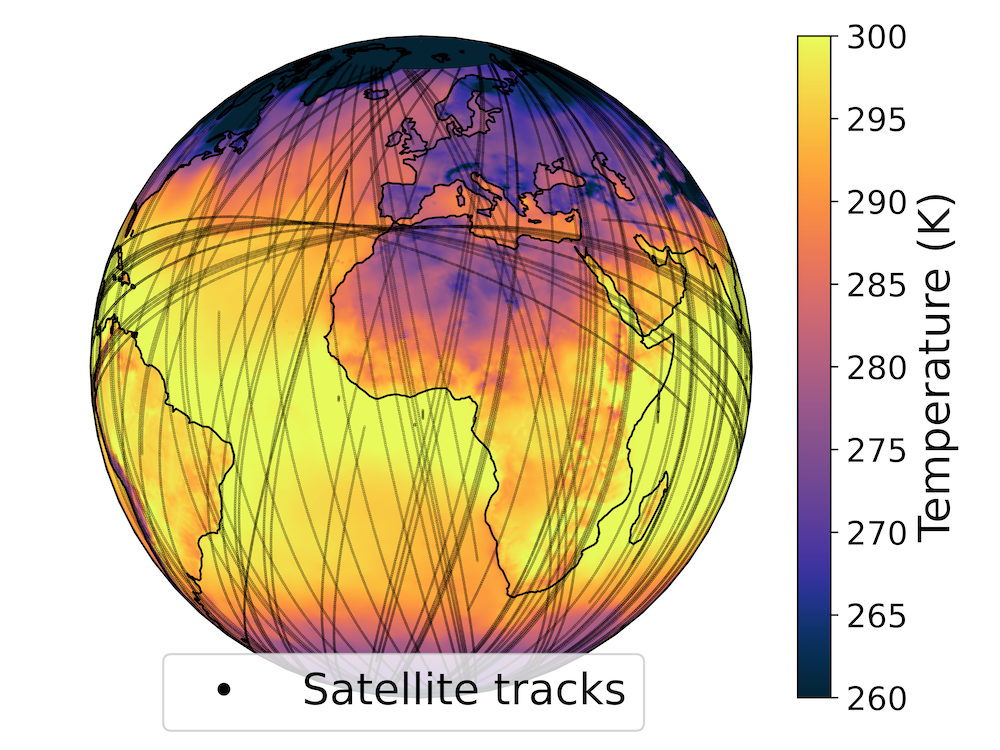}
    \caption{
        Surface temperature computed by message passing from satellite observations.
        The lines show the locations of the observations.
    }
    \label{fig:temperature_data_ours}
\end{wrapfigure}
Data assimilation (DA) is a core component of numerical weather prediction (NWP) systems. The goal of DA is to provide the best estimate of the current state of the dynamical system (the atmosphere in NWP). This is achieved by combining the forecasted state of the system with measurements, for example satellites (tracks of which can be seen in Figure \ref{fig:temperature_data_ours}), sensors on the ground and weather balloons. A corrected forecast is then produced using the resultant estimate as the initial condition. Due to the number of observations and dimensionality of the state, DA consumes a large amount of computation time and memory.

In recent years, the scalability of DA approaches has been pushed to the limit, with operational weather centres launching programmes to solve the problem \parencite{ecmwf2020scalability}. Several challenges are identified.
First, the amount of data that must be processed has grown dramatically due to the increasing availability of satellite observations and the higher resolution of forecasting models.
Second, while the total computational power available continues to increase, this is no longer due to individual cores getting faster but instead in the form of expanded parallelism.
Thus, DA algorithms must be able to take full advantage of parallel and distributed hardware.

4D-Var and 3D-Var, and their derivatives, are the DA algorithms used by many systems for weather forecasting and other applications, such as ocean simulation \parencite{bonavita2021, saulter2020}.
A standard approach to distribute these algorithms across many compute nodes is spatial parallelism using domain decomposition \parencite{damore2014scalable,arcucci2015parallel}.
The geographic area covered by the model is divided into overlapping subdomains; each subdomain is assigned to a single compute node that solves the assimilation problem in that subdomain.
A key limitation of this approach is that the overlapping regions between subdomains must be carefully synchronised between nodes to ensure that the states computed on each node are physically consistent with each other.
These synchronisation steps introduce a bottleneck due to communication overheads, and this can be exacerbated by any computing load imbalance between the subdomains.
Although there have been attempts to relax the level of synchronisation required \parencite{cipollone2020parallel}, ideally synchronisation requirements would be removed altogether.

In this paper, we propose an alternative approach to DA that is designed with distributed computation in mind.
We exploit the formulation of DA as a Bayesian inference problem \parencite{Evensen2022}, which allows us to apply tools from the large-scale Bayesian inference literature.
In particular, we develop a method based on considering DA as inference in a Gaussian Markov random field (GMRF) and develop a message-passing algorithm to perform inference on this field.
This approach naturally supports domain decomposition across multiple nodes \emph{without} requiring overlapping regions and, therefore, no synchronisation is required.
Only a small amount of data must be communicated between subdomains, and this can be performed asynchronously with the computation on each node.
In this initial work, we consider only two-dimensional spatial inference problems, rather than the three-dimensional or three-dimensional-with-time problems commonly seen in applications. However, the framework we develop here is designed to be extended to the full three-dimensional-with-time case.
The contributions of this work are as follows:
\begin{itemize}
    \item We propose an approach for expressing the DA problem as a message-passing algorithm.
    \item We develop a GPU-accelerated implementation for maximum a posteriori inference, which naturally supports distributed computation for very large domains.
    \item We demonstrate the efficacy of our algorithm on surface temperature data and demonstrate that our approach is a viable DA technique.
    \item We also include a fast, GPU-accelerated implementation of 3D-Var as a very strong baseline.
\end{itemize}
We release our message passing and 3D-Var implementations, and code to reproduce our experiments, at
\href{https://github.com/oscarkey/message-passing-for-da}{github.com/oscarkey/message-passing-for-da}.

\section{Background}
In this section, we recall the Bayesian formulation of data assimilation and discuss several inference algorithms.
In particular, we introduce the message-passing algorithm that we apply in this work.

\subsection{Data Assimilation as Bayesian Inference}
Data assimilation (DA) comprises a set of techniques in numerical weather prediction that aim to combine earth observations with assumptions about the state of the weather to produce an updated estimate of the weather.
In this work, we simplify the full DA problem to spatial inference of a single variable.
Let $\vec{f} = (f_1, \ldots, f_n) \in \Reals^{n}$ be a weather variable, such as surface temperature, that is discretised on a large 2D spatial grid consisting of points $\vec{x}_1,\ldots,\vec{x}_n$.
We also have $m$ observations, $\vec{y} \in \Reals^{m}$, which may either be derived from remote sensing products or direct observations of atmospheric variables.
From a probabilistic perspective, DA can then be understood as a Bayesian inference problem: our assumptions about the weather can be encoded as a prior $p(\vec{f})$, the observations as the likelihood $p(\vec{y} | \vec{f})$, and our updated estimate as the posterior computed via Bayes' theorem as
$p(\vec{f} | \vec{y}) \propto p(\vec{y} | \vec{f}) p(\vec{f})$.
$n$ is typically in the billions and $m$ in the tens of millions \parencite{metofficeDA}, making direct inference using Bayes' theorem impossible.
Thus, the choice of prior and likelihood is heavily influenced by the need to make inference efficient, and we discuss several options in the next section.

In operational DA systems, the problem is more complex, consisting of a 3D spatio-temporal grid and multiple weather variables at each grid point.
However, inference in the simplified setting above is still challenging for large $n$ and $m$.
In \cref{sec:conclusion} we discuss extensions to the full DA problem.

\subsection{Existing Methods for Large-Scale Inference}

\paragraph{Optimal Interpolation}
Optimal interpolation \parencite[Section 5.4.1]{kalnay2003atmospheric}, is virtually synonymous with Gaussian process (GP) regression from the machine learning literature \parencite{williams2006gaussian}.
This is the basis of all the DA methods discussed in this work, with subsequent methods being approximations to it.
GP regression assumes a Gaussian prior $p(\vec{f}) = \mathcal N\left( \vec f_b, \mat\Sigma \right)$ and likelihood $p(\vec{y} | \vec{f}) = \mathcal N\left( \mat{H} \vec{f}, \mat{R} \right)$.
Here $\vec{f}_b \in \Reals^{n}$ is the prior mean, and $\mat\Sigma \in \Reals^{n \times n}$ the prior covariance defined by a function $k$ (known as a reproducing kernel), where $\Sigma_{i,j} = k(\vec{x}_i, \vec{x}_j)$.
$\mat{H} \in \Reals^{m \times n}$ and $\mat{R} \in \Reals^{m \times m}$ are the linear observation operator and diagonal error covariance, which are either determined using prior knowledge or learned from data.
Under this prior and likelihood, the posterior $p(\vec{f} | \vec{y})$ is also a Gaussian, having a closed form expression.
Unfortunately, computing the desired posterior costs $\mathcal{O}(l^3 + nl)$, where $l$ is the number of grid points at which there are observations.
A modern solution to large-scale inference with GPs is to use inducing-point methods, which approximate the observations with a much smaller number of pseudo data points \parencite{titsias2009variational}. However, the estimates obtained may be too crude for practical use at the scale that is typically considered in numerical weather forecasting.

\paragraph{Gaussian Markov Random Fields}
Another solution to reducing the cubic cost is to use a prior $p(\vec{f})$ defined by a {\em Gaussian Markov random field (GMRF)} \parencite{rue2005gaussian}.
This approach makes use of the spatial interpretation of $\vec{f}$ to make a Markovian assumption that each $f_i$ only directly depends on other $f_j$ in its neighbourhood.
Additionally, the prior $p(\vec{f})$ and inference process are expressed in terms of the inverse of the covariance matrix, known as the precision.
Under the Markovian assumption, the precision matrix is sparse, allowing inference in GMRFs to scale $\mathcal{O}(n^{3/2})$ in 2D and $\mathcal{O}(n^{2})$ in 3D (the complexity does not depend on the number of observations).
In addition, the INLA framework \parencite{rue2009approximate} makes it possible to handle nonlinear observation models and infer the model hyperparameters from data, although this requires further approximations.
A downside of the approach is that it is inherently sequential. Thus, it cannot make effective use of modern parallel computing hardware, such as GPUs, or be distributed across several compute nodes.

\paragraph{3D-Var}
Another alternative that reduces the cost of GP regression is to only compute the maximum a posteriori (MAP) estimate, rather than the full posterior, that is find $\vec{f}_\text{MAP} = \argmax_{\vec{f}} p(\vec{f} | \vec{y})$.
This is the approach taken by 3D-Var, which is known as a ``variational'' method.
The maximisation can also be expressed as minimising the cost function $J[\vec{f}] = -\log{p(\vec{f} | \vec{y})} = -\log{p(\vec{y}|\vec{f}}) - \log{p(\vec{f})} + C$, where $C$ is a constant that does not depend on $\vec{f}$.
Substituting in the GP prior and Gaussian likelihood, the cost function becomes
\begin{equation}\label{eq:3dvar-cost}
    J[\vec{f}] = \frac12 (\vec{y} - \mat{H}(\vec{f}))^\top \mat{R}^{-1}(\vec{y} - \mat{H}(\vec{f})) + \frac12 (\vec{f} - \vec{f}_b)^\top \mat{\Sigma}^{-1} (\vec{f} - \vec{f}_b).
\end{equation}
In practice, this is minimised using an optimiser, for example, L-BFGS \parencite{liu1989lbfgs} (as seen in \textcite{damore2015lbfgs}) or a Krylov solver (as seen in \textcite{ecmwf2020scalability}).
3D-Var is more flexible than optimal interpolation, as it can handle nonlinear observation models $\mat{H}$. However, it has the downside of not being able to provide uncertainty estimates, as it is a MAP estimator.
In weather forecasting applications, 3D-Var is usually extended to 4D-Var, which includes a time component.

\subsection{Factor Graphs and Inference with Message Passing} \label{sec:message_passing_background}
In this work, we perform inference using message passing \parencite{Kschischang2001}, which computes the marginals of any joint probability model that can be expressed as a factor graph.
We introduce message passing for a general continuous distribution $g(\vec{f}) = g(f_1, \ldots, f_n)$, and specialise it to our posterior in \cref{sec:message_passing_method}.
$g(\vec{f})$ can be expressed as a factor graph if it has a known decomposition
\begin{equation}\label{eq:joint-prob-model}
    g(\vec{f}) \propto \prod_{i = 1}^n \phi_i(f_i) \prod_{j = 1}^n \phi_{ij}(f_i, f_j),
\end{equation}
where $\phi_i$ and $\phi_{i,j}$, referred to as the {\em nodewise} and {\em pairwise factors}, respectively, are functions from a variable, or a pair of variables, to $\Reals$. These do not have to be probability distributions.
The factorisation in \eqref{eq:joint-prob-model} induces a sparse graph on the variables $\{f_i\}_{i=1}^n$, where two nodes $f_i$ and $f_j$ are connected if and only if $\phi_{i,j}$ is non-constant.
Given such a graph, the algorithm associates a pair of \emph{messages} on each edge at each iteration $t$:
a message $m^t_{ij} = (a^t_{ij},b^t_{ij}) \in \Reals \times \Reals$ from $f_i$ to $f_j$, and a similar message $m^t_{ji}$ from $f_j$ to $f_i$.
We use the message passing variant introduced by \textcite{Ruozzi2013} (in turn a generalisation of \textcite{wiegerinck2002fractional}).
This is summarised in \cref{alg:message_passing} and illustrated for our application in \cref{fig:message_passing_intuition}, and \Cref{app:message_passing_background} describes it in more detail.
\begin{algorithm}[H]
\caption{Re-weighted message passing}
\label{alg:message_passing}
\begin{algorithmic}[1]
    \Procedure{Message Passing}{$\{f_i\}_{i=1}^n$, $\{\phi_i\}_{i=1}^n$, $\{\phi_{ij}\}_{i,j=1}^n$}
    \Comment input is factor graph
    \State $m^{t=0}_{ij} = (0, 10^{-8})$ for all $i$, $j$
    \For{$t \in \{1, \ldots T\}$}
        \For{$f_i$ in the graph}
            \For{$f_j \in \{f_j \sim f_i \}$}
                \State $m^t_{ij} = \Call{Compute Outgoing Message}{
                    c,
                    \phi_i,
                    \{ (\phi_{ki}, m^{t-1}_{ki}) : \{f_k \sim f_i \} \backslash f_j \}
                }$
                \label{algline:message_update}
            \EndFor
        \EndFor
    \EndFor
    \State \Return $g(f_i) = \Call{Compute Marginal}{
                        c, \{m^T_{ki} : f_k \sim f_i \}
                    }$
            for all $i$
    \EndProcedure
\end{algorithmic}
\end{algorithm}
We use the notation $\{f_k \sim f_i\}$ to denote the set of all variables $f_k$ that are connected with $f_i$,
$\Call{Compute Outgoing Message}$ and $\Call{Compute Marginal}$ are defined formally in \cref{app:full_algorithm},
$T$ is the total number of iterations, and $c \in \Reals$ is a hyperparameter to re-weight contributions of the messages, necessary to aid convergence.
Note that each iteration of the inner \texttt{for} loops is independent, and each node only writes and reads messages with the nodes directly connected to it. The algorithm is therefore very amenable to distributed computation.

While this instance of message passing can theoretically compute both the mean and variance of the marginals, in practice the variance estimates are biased and do not provide useful estimates of the uncertainty \parencite{weiss1999correctness}.
Thus, we only use message passing to compute the posterior mean, the MAP estimate of the posterior.
This makes message passing an alternative to 3D-Var.

\section{Data Assimilation with Message Passing}
Our method begins by placing a Mat\'ern GP prior over the domain, which is the de facto standard model choice in spatial geostatistics \parencite{guttorp2006studies}, and assuming a Gaussian likelihood.
We discretise the prior to a GMRF and derive the corresponding factor graph.
Then, we apply a message-passing algorithm to the graph and the observations to compute the marginal posterior means.

\subsection{Derivation of the Factor Graph} \label{sec:factor_graph_derivation}
The Mat\'ern GP prior can be characterised as the solution to a stochastic partial differential equation (SPDE) of the form
\begin{equation}\label{eq:matern-spde}
    (\kappa^2 - \Delta)^{\alpha/2} f = \mathcal{W},
\end{equation}
where $f$ is the process, $\Delta$ is the Laplacian operator, $\kappa$, $\alpha$ are positive hyperparameters and $\mathcal{W}$ is the spatial white-noise process with spectral density $\sigma^2 q$, for hyperparameters $\sigma,q \in \Reals$.
Following \textcite{lindgren2011explicit}, we first derive a GMRF representation of the Mat\'ern GP by discretising this SPDE using finite differences (finite elements would also be possible). On a uniform 2D grid with step sizes $\Delta x$ and $\Delta y$ in the $x$ and $y$-directions respectively, this yields a random matrix-vector system
\begin{equation}\label{eq:discretised-spde}
    \mat{L} \vec{f} = \vec{w}, \quad \text{where} \quad \vec{w} = \sqrt{\frac{\sigma^2 q}{\Delta x \Delta y}} \vec{z}, \quad \vec{z} \in \Reals^n \sim \mathcal{N}(\vec{0}, \mat{I}_n).
\end{equation}
Here, $\mat{L} \in \Reals^{n \times n}$ is a matrix representing the operator $\mathcal{L} := (\kappa^2 - \Delta)^{\alpha/2}$ under discretisation, which is guaranteed to be sparse if the exponent $\alpha/2$ is an integer \parencite{lindgren2011explicit}, and $\vec{f}, \vec{w}$ are finite-dimensional vector representations of the random fields $f$ and $\mathcal{W}$.
(see Appendix \ref{app:discretisation} for details on the discretisation). Now, \eqref{eq:discretised-spde} implies that $\vec{f}$ is a Gaussian random variable of the form
\begin{align}
    \vec{f} \sim \mathcal{N}(\vec 0, (\gamma \mat{L}^\top \mat{L})^{-1}), \quad \text{where} \quad \gamma := \frac{\Delta x \Delta y}{\sigma^2 q},
\end{align}
which is a GMRF, since its precision matrix $\mat{P} := \gamma \mat{L}^\top \mat{L}$ is sparse (here, $\mat{L}$ is sparse and banded).
We can build a graph from this GMRF by the following simple rule: Take all of $f_1, \ldots, f_n$ as nodes in the graph and connect two nodes $f_i$ and $f_j$ by an edge if $[\mat{P}]_{ij} \neq 0$. Then, we have
\begin{align}
    p(\vec{f}) \propto \exp\left(-\frac12 \vec{f}^\top \mat{P} \vec{f}\right) = \exp\left(-\sum_{i=1}^n \frac12 [\mat{P}]_{ii} f_i^2 - \sum_{j \sim i} [\mat{P}]_{ij} f_i f_j \right),
\end{align}
where $\Sigma_{j \sim i}(\cdot)$ denotes the sum over all indices $j$ that are adjacent to $i$ in the graph. Setting
\begin{align}\label{eq:factors}
    \phi_i(f_i) := \exp\left(-\frac12 [\mat{P}]_{ii} f_i^2\right) \quad \text{and} \quad \phi_{ij}(f_i, f_j) := \exp\left(-[\mat{P}]_{ij} f_i f_j\right),
\end{align}
we have that the prior $p(\vec{f}) \propto \prod_{i = 1}^N \phi_i(f_i) \prod_{j \sim i}\phi_{ij}(f_i, f_j)$, thus we have a factor graph representation.

\begin{figure}
    \begin{minipage}[t]{0.45\textwidth}
        \centering
        \includegraphics[height=4cm]{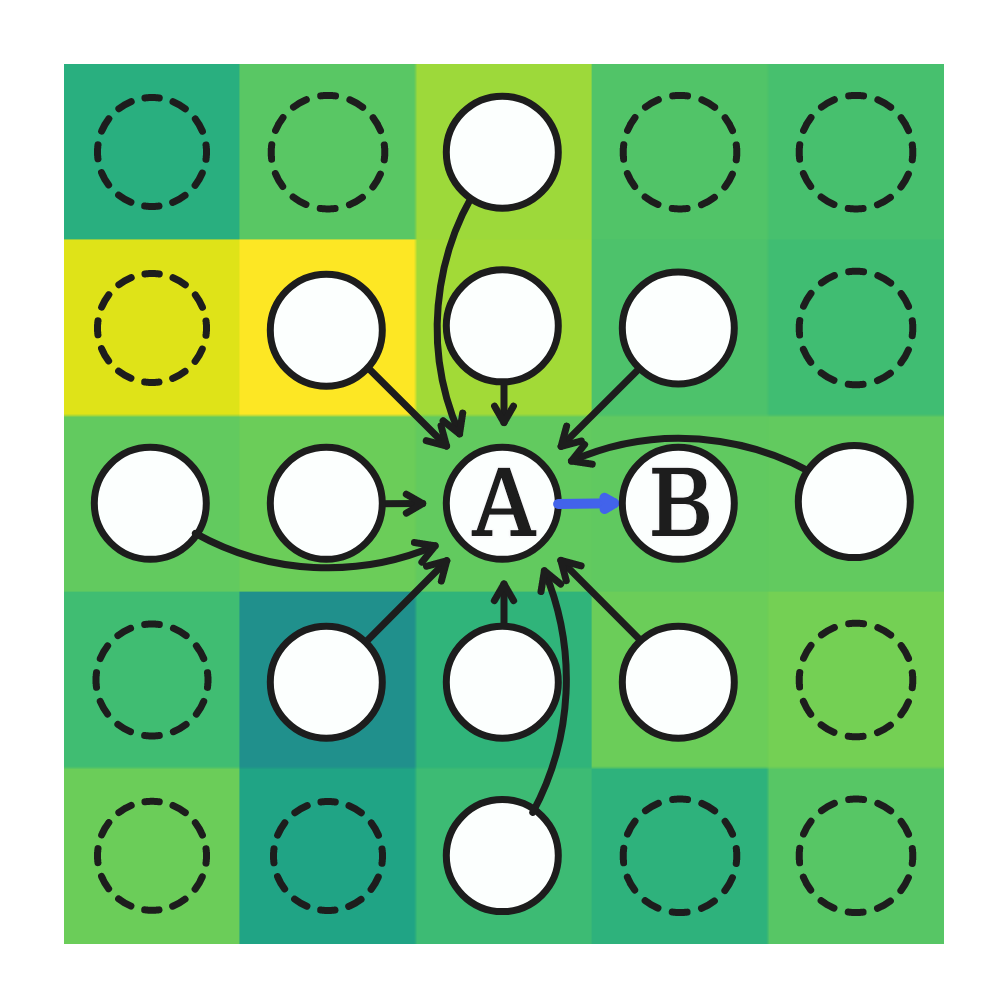}
        \caption{
            Illustration of node A sending a message to node B. Black arrows indicate incoming messages that are combined to compute the outgoing message in blue.
        }
        \label{fig:message_passing_intuition}
    \end{minipage}\hfill
    \begin{minipage}[t]{0.49\textwidth}
        \centering
        \includegraphics[width=7cm,height=4cm]{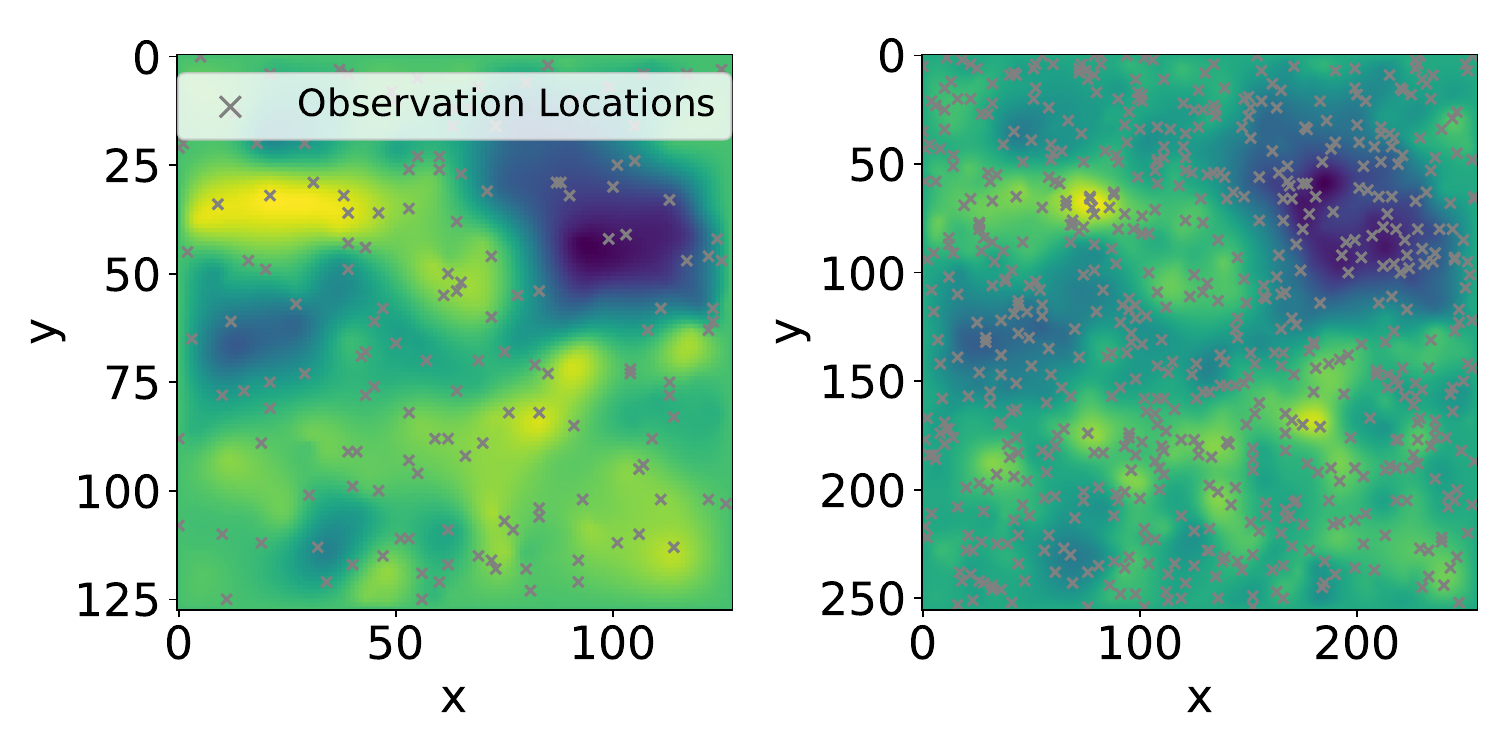}
        \caption{Illustration of the multigrid implementation, showing the marginal means computed at two levels ($128\times 128$ to $256 \times 256$) of resolution on the simulated data.}
        \label{fig:multigrid_demo}
    \end{minipage}
\end{figure}

\subsection{Computing the Posterior Mean with Message Passing} \label{sec:message_passing_method}
Having calculated the factor graph corresponding to the prior, we can now apply message passing (\cref{alg:message_passing}) to combine this with the observations and compute the posterior marginals.
We make several modifications to tailor the algorithm to DA, which we summarise below and detail in \cref{app:full_algorithm}.

\paragraph{Including Observations}
To add information about the observations $\vec{y}$ into our factor graph, we modify the nodewise factor in \eqref{eq:factors} by $\phi_i(f_i) \mapsto p(y_i | f_i) \phi_i(f_i)$.
In our experiments, we assume that the weather variable is noisely observed at a subset of grid cells, where the noise $\sigma_y^2 \in \Reals$ is constant for all observations.
Thus, we set $p(y_i | f_i) = \mathcal{N}(y_i | f_i, \sigma_y^2)$ at grid points $\vec{x}_i$ where there is an observation, and $p(y_i | f_i) = \mathcal{N}(y_i | 0, z)$, for very large $z$, where there is not.

\paragraph{Update Damping}
To improve the stability of the algorithm, we follow \textcite{pretti2005damping} and dampen the updates of the messages, replacing line~\ref{algline:message_update} of \cref{alg:message_passing} with
\begin{equation*}
    m^{t+1}_{ij} = (1 - \eta) m^{t} + \eta \, \Call{Compute Outgoing Message}{\cdot},
\end{equation*}
where $\eta \in (0,1)$ is a hyperparameter, which we refer to as the learning rate.

\paragraph{Early topping}
To avoid specifying the total number of iterations as a hyperparameter, we choose a generic large $T$ and stop when the change in message between iterations is smaller than a threshold.

\paragraph{Multigrid}
We apply a multigrid technique to speed up the convergence of the algorithm.
These have been used extensively when solving partial differential equations numerically, and provide an efficient way of accelerating the convergence of iterative approaches if multiscale phenomena are being modelled, with grids at different resolutions capturing different spatial scales.
In the case of message-passing, we can intuitively view information propagating from the observation locations across the graph.
If the density of the observations is low, this can take many iterations.
To solve this the multigrid approach starts with a low resolution grid, and iterates message passing until convergence---this is fast on a low-resolution grid.
It then doubles the size of the grid, initialising the messages to the converged messages from the previous grid. This process is repeated until we reach the target grid size. Observations are taken on the target resolution and introduced at each multigrid level when the observation location exactly coincides with the grid level coordinates.
\Cref{fig:multigrid_demo} illustrates the procedure.

\subsubsection{Computational Efficiency}
The Markovian assumption made by the GMRF prior from which we derive the factor graph results in a sparse graph in which each node is only connected to nodes in its local area.
The connectivity is also very regular, with each node connected to the same set of relative nodes except at the edges of the graph.
These two properties make the graph high amenable to GPU computation, using an approach similar to \parencite{zhou2022pgmax} but optimised for our particular graph structure.

The main advantage of the message-passing approach is that it can naturally be distributed by dividing the nodes of the graph into subdomains and splitting them between compute nodes.
The only communication required between compute nodes is to update the messages on the borders of the subdomains, which is a small amount of data.
Additionally, there is no requirement that the border messages are updated every iteration, so they could be updated asynchronously to the computation within each subdomain.

\section{Experiments}
We evaluate the performance of message passing on both simulated data and a more realistic surface temperature DA problem.
We compare against two baselines: the GMRF method, using the R-INLA implementation \parencite{lindgren2015bayesian}, and 3D-Var.
Note that R-INLA computes the exact marginals of the posterior, while 3D-Var and message passing are approximate methods that compute only the mean.
Thus, R-INLA provides a reference for the best-case error that the approximate methods could achieve.

\paragraph{GPU-accelerated 3D-Var Implementation}
The 3D-Var cost function is obtained by substituting in the same GMRF prior and Gaussian likelihood as used for message passing, and then minimised using the L-BFGS optimiser.
We use our own GPU-accelerated implementation using the experimental support for sparse linear algebra in JAX \parencite{jax2018} and JAXopt \parencite{jaxopt_implicit_diff}.
We expect this implementation to be significantly faster than the CPU implementations currently in deployment.
We also implement message passing in JAX with GPU acceleration, while R-INLA runs on the CPU only.

\paragraph{Hyperparameters}
We perform a grid search, reported in \cref{app:grid_search}, to select the message weighting and learning rate hyperparameters of message passing, and the early stopping thresholds for both message passing and 3D-Var.
We note that selecting the message weighting and learning rate is quite easy: the algorithm converges for a wide range of choices and, if it does converge, the exact choice has only a small effect on the speed of convergence.
\Cref{app:experiment_details} gives the remaining details of all experiments.

\subsection{Effect of Domain Size and Observation Density}\label{sec:syntheitc-experiment}
Our first set of experiments are performed on simulated data.
We sample a square ground truth field from a GMRF prior, randomly select a given fraction of the grid points as observations, and perform inference against these observations and the same prior.
\Cref{tab:comparison} shows the RMSE between the posterior mean and the ground truth and the time taken, as we vary the grid size and the observation density.
In the message passing runs the multigrid approach is used, with a base grid size of $32 \times 32$ in all cases.

The results show that 3D-Var and message passing achieve similar RMSEs in all cases, although both have more error than R-INLA when only 1\% of the grid is observed.
When 5\% and 10\% of the grid is observed, 3D-Var and message passing take similar amounts of time; however, message passing is significantly slower for a larger grid when only 1\% is observed.
While the iteration time of message passing is independent of the number of observations, as the observation density falls it takes an increasing number of iterations for the information to propagate from the observed points across the grid, thus early stopping happens later.
R-INLA is much slower than the other methods, because it is a sequential method that cannot take advantage of the GPU.

\begin{table}[h]
\caption{
    Comparison on simulated data.
    We give the mean over three ground truths; we do not observe significant variance (therefore omitted).
    Bold indicates where either 3D-Var or message passing performed better.
    R-INLA is included to show the minimal achievable error given the prior, as it computes an exact posterior.
}
\label{tab:comparison}
    \centering
\begin{tabular}{rr|rrr|rrr}
 &  & \multicolumn{3}{|c|}{\textbf{RMSE}} & \multicolumn{3}{c}{\textbf{duration (seconds)}} \\
\hline
\textbf{grid size} & \textbf{observations} & R-INLA & 3D-Var & MP & R-INLA & 3D-Var & MP \\
\hline
\multirow[t]{3}{*}{$256 \times 256$} & 1\% & 0.192 & \textbf{0.202} & 0.213 & 19.4 & \textbf{3.6} & 7.6 \\
 & 5\% & 0.093 & 0.093 & 0.093 & 21.6 & 4.1 & \textbf{1.7} \\
 & 10\% & 0.069 & 0.069 & \textbf{0.068} & 22.5 & 4.5 & \textbf{0.9} \\
\cline{1-8}
\multirow[t]{3}{*}{$512 \times 512$} & 1\% & 0.101 & 0.128 & \textbf{0.127} & 107.9 & \textbf{3.9} & 15.9 \\
 & 5\% & 0.047 & 0.048 & 0.048 & 104.8 & 5.3 & \textbf{4.2} \\
 & 10\% & 0.034 & 0.036 & \textbf{0.034} & 99.8 & 5.9 & \textbf{2.5} \\
\cline{1-8}
\multirow[t]{3}{*}{$1024 \times 1024$} & 1\% & 0.050 & 0.116 & \textbf{0.066} & 601.6 & \textbf{4.5} & 50.5 \\
 & 5\% & 0.024 & 0.026 & \textbf{0.024} & 848.7 & \textbf{8.0} & 13.8 \\
 & 10\% & 0.017 & 0.020 & \textbf{0.017} & 547.5 & 11.5 & \textbf{8.7} \\
\end{tabular}
\end{table}

\subsection{Large-scale Example}
For a more realistic use case, we consider the global surface temperature field.
We take the ground truth data from a run of the Met Office's Unified Model \parencite{UM2019} at N1280 resolution, where the data is valid for 06UTC 2020-01-01.
To avoid issues with boundary conditions, we consider a clipped domain with dimensions $2500\times1500 = 3.75M$ grid points.
We use spherical polar coordinates.
The observation locations are generated from the geographical positions (latitude, longitude) of weather-focused satellites calculated over a 3-hour window, which corresponds to $\approx8\%$ of the grid being observed.
As the prior mean we select a climatology mean of the global surface temperature calculated from ERA5 \parencite{era2020}.

\Cref{fig:temperature_data_ours} showcases the resultant mean estimates from the message passing approach, while error plots are shown in \cref{fig:temperature_data_l1} in \cref{app:additional_results}.
3D-Var achieved an area-weighted RMSE of $2.33$ K and took $16$ seconds, while message passing achieved an area-weighted RMSE of $1.23$ K and took $115$ seconds. For comparison, the area-weighted RMSE calculated for the prior mean (ERA5) against the high-resolution temperature field is $2.78$ K.
R-INLA did not complete after $>1$ hour of processing time; thus, we do not include its results.

\section{Conclusion} \label{sec:conclusion}
In this paper, we present a new perspective on DA based on insights from the literature on large-scale Bayesian inference.
We demonstrate that our message-passing approach is viable and can, in many scenarios, produce results competitive with a GPU-accelerated 3D-Var implementation.
In the era of large heterogeneous computing systems, the scalability issues with state-of-the-art variational DA are well documented and the design of the proposed method should offer improved scalability.
However, further research is required to determine if this design offers advantages for operational-scale problems.

We make several simplifying assumptions.
First, we have only considered spatial inference in two dimensions. However, our approach can be extended to full spatio-temporal inference as we describe in \cref{app:spatiotemporal_extension}.
We have also only considered a single weather variable and linear observation operators. However, this can be extended relatively easily to multiple weather variables under the current framework, by assuming that they are independent under the prior. Further work is required for variables coupled in the prior.
It may also be possible to support non-linear observation operators using iterative linearisation techniques \parencite{kamthe2022iterative}.

The primary limitation of our approach is that message passing only reliably computes the posterior mean; the obtained posterior marginal uncertainties are inherently biased when the graph is loopy \parencite{weiss1999correctness}. Thus, we cannot get an accurate estimate of the uncertainty in the assimilated state.
It also requires Gaussianity assumption in the prior, although for non-Gaussian priors arising from nonlinear stochastic PDEs, we may be able to handle this using an iterative linearisation method \parencite{anderka2024iterated}. Finally, due to our unreliable uncertainty estimates, we cannot make use of the marginal likelihood to learn the hyperparameters of the prior, for example, the kernel lengthscale. Currently, we handle this using cross-validation on held-out observations.
We note however that all of these limitations are shared with popular large scale DA methods such as 3D-Var and resolving these issues will be a significant step forward for future DA research.

\FloatBarrier

\begin{Backmatter}

\paragraph{Acknowledgments}
We are grateful for the technical assistance of Dr. Cyril Morcrette for providing the Met Office's Unified Model high-resolution temperature data. We are also grateful for Prof. Mohammad Emtiyaz Khan and Prof. Nicholas Ruozzi for useful discussions and feedbacks.

\paragraph{Provenance}
This article was accepted into the Climate Informatics 2024 (CI2024) Conference.
It has been published in Environmental Data Science on the strength of the CI2024 review process.

\paragraph{Funding Statement}
OK acknowledges support from the Engineering and Physical Sciences Research Council with grant number EP/S021566/1. ST is supported by a Department of Defense Vannevar Bush Faculty Fellowship held by Prof. Andrew Stuart,
and by the SciAI Center, funded by the Office of Naval Research (ONR), under Grant Number N00014-23-1-2729.

\paragraph{Competing Interests}
The authors declare none.

\paragraph{Data Availability Statement}
Our message passing and 3D-Var implementations, and code to reproduce our experiments, is available at \url{https://github.com/oscarkey/message-passing-for-da}.
It is also archived with at \url{https://doi.org/10.5281/zenodo.13137965}.
The data for the surface temperature data experiments is taken from the Met Office's Unified Model, and thus sadly cannot be publicly released.
The data for the other experiments is generated automatically by the experiments.

\paragraph{Ethical Standards}
The research meets all ethical guidelines, including adherence to the legal requirements of the study country.

\paragraph{Author Contributions}
Conceptualisation: ST, MD; Methodology: ST, MD;  Software: OK, DG; Data curation: DG; Data visualisation: DG, OK; Writing - original draft: OK; Writing - Review and Editing: OK, ST, DG, MD; Supervision: ST, MD. All authors approved the final submitted draft.

\paragraph{Supplementary Material}
Appendices included in this document.

\end{Backmatter}

\pagebreak

\begin{appendix}\appheader

\section{Background on Message Passing for Bayesian Inference} \label{app:message_passing_background}
In this section, we provide a formal description of message-passing algorithms for Bayesian inference.
Then, in \cref{app:full_algorithm}, we give full details of the particular message-passing algorithm we use in this work.

Given a factorisation of the joint probability model $g(\vec{f}) = g(f_1, \ldots, f_n)$ of the form
\begin{equation}
    g(\vec{f}) \propto \prod_{i = 1}^n \phi_i(f_i) \prod_{j = 1}^n \phi_{ij}(f_i, f_j),
\end{equation}
for nodewise and pairwise factors $\phi_i$ and $\phi_{i,j}$, this induces a sparse graphical structure on the variables $\{f_i\}_{i=1}^n$, where two nodes $f_i$ and $f_j$ are connected if and only if the pairwise factor $\phi_{i,j}$ is non-constant.
We can further augment this graph by including the non-constant factors $\{\phi_i\}_{i \in V}$ and $\{\phi_{ij}\}_{(i,j)\in E}$ as additional nodes, and joining the pairs of nodes $(\phi_i, f_i)$, $(\phi_{ij}, f_i)$, and $(\phi_{ij}, f_j)$ by edges. This augmented structure is referred to as the {\em factor graph representation} of the probabilistic model \eqref{eq:joint-prob-model}, which is a bipartite graph with one side of nodes being the set of variables $f_i$ and the other side being the set of factors $\phi_i, \phi_{ij}$.

\subsection{Loopy Belief Propagation}
Given a factor graph representation of the model, one can approximate the marginal probabilities $\{g(f_i)\}_{i=1}^n$ efficiently by a message-passing algorithm, referred to as {\em loopy belief propagation}. This begins by defining a set of so-called \emph{messages} $m^t_{f_i \rightarrow \phi_\alpha}(f_i)$, $m^t_{\phi_\alpha \rightarrow f_i}(f_i)$ between each of the variable and factor nodes.
The messages are initialised to $m^0_{f_i \rightarrow \phi_\alpha}(f_i) = m^0_{\phi_\alpha \rightarrow f_i}(f_i) = 1$ and updated by iterating the following steps:
\begin{align}
    m^{t+1}_{f_i \rightarrow \phi_\alpha}(f_i) &= \phi_i(f_i) \prod_{\substack{\phi_\beta \sim f_i \\ \phi_\beta \neq \phi_\alpha}} m^t_{\phi_\beta \rightarrow f_i}(f_i), \quad i=1, \ldots, n, \label{eq:variable-to-factor-message} \\
    m^{t+1}_{\phi_\alpha \rightarrow f_i}(f_i) &= \int_{\mathbb{R}} \phi_{ij}(f_i, f_j) m^t_{f_{j} \rightarrow \phi_\alpha}(f_{j}) \,\mathrm{d} f_j, \quad i=1, \ldots, n. \label{eq:factor-to-variable-message}
\end{align}
The updates \eqref{eq:variable-to-factor-message}--\eqref{eq:factor-to-variable-message} can be done either in parallel, serially, or following more specific scheduling rules \parencite{elidan2006residual, gonzalez2009residual, van2019message} -- generally, there are no restrictions on the order in which we pass the messages between nodes.
Upon convergence, say at iteration $t=T$, we can then estimate the marginal distribution at each variable $g(f_i)$, by taking
\begin{align}\label{eq:state-update}
    g(f_i) \approx \frac{1}{Z}\prod_{\phi_\alpha \sim f_i} m^T_{\phi_\alpha \rightarrow f_i}(f_i), \quad \text{where} \quad Z = \prod_{\phi_\alpha \sim f_i} \int_{\mathbb{R}} m^T_{\phi_\alpha \rightarrow f_i}(f_i) \mathrm{d} f_i,
\end{align}
for $i=1, \ldots, n$.
It is well-known that, when the graph is tree-structured, convergence is guaranteed and the message-passing algorithm yields the exact marginals.
However, if the graph contains loops, then convergence is not guaranteed and moreover the obtained marginals may not be exact.
Various techniques have been proposed to improve this. In particular, one can make {\em fractional updates} to the messages \parencite{wiegerinck2002fractional}, which we express in the form
\begin{align}
    m_{f_i \rightarrow \phi_\alpha}(f_i)^{\frac{1}{c}} &= \phi_i(x_i) \prod_{\substack{\phi_\beta \sim f_i \\ \phi_\beta \neq \phi_\alpha}} m_{\phi_\beta \rightarrow f_i}(f_i) m_{\phi_\alpha \rightarrow f_i}(f_i)^{1-\frac1c}, \label{eq:reweighted-variable-to-factor-message} \\
    m_{\phi_\alpha \rightarrow f_i}(f_i)^{\frac{1}{c}} &= \int_{\mathbb{R}} \phi_{ij}(f_i, f_j)^{\frac{1}{c}} m_{f_{j} \rightarrow \phi_\alpha}(f_{j})^{\frac1c} \,\mathrm{d} f_j, \label{eq:reweighted-factor-to-variable-message}
\end{align}
for constants $c_\alpha \in \mathbb{N}$.
In this work, we use an extension of this by \textcite{Ruozzi2013}, where $c_\alpha$ is also allowed to take any real values, including negative ones. A benefit of this re-weighting scheme is that one can guarantee convergence even in the loopy setting by taking $c_\alpha$ large enough, or by setting them to be negative \parencite{Ruozzi2013}. While the message updates in the re-weighted scheme are modified according to \eqref{eq:reweighted-variable-to-factor-message}--\eqref{eq:reweighted-factor-to-variable-message}, the marginal computation is still performed in the same way as standard message passing \eqref{eq:state-update}.

Next, we discuss the details on how to compute the messages \eqref{eq:variable-to-factor-message}--\eqref{eq:factor-to-variable-message} and \eqref{eq:reweighted-variable-to-factor-message}--\eqref{eq:reweighted-factor-to-variable-message} more concretely in the special but important case of joint Gaussian probability models.

\subsection{Gaussian Setting}
For Gaussian probability models, which we assume to have zero mean for the time being, we can write down the joint probability explicitly as
\begin{align}
    p(\vec{f}) \propto \exp\left(-\frac12 \vec{f}^\top \mat{P} \vec{f}\right) = \exp\left(-\sum_{i \in V} \frac12 [\mat{P}]_{ii} f_i^2 - \sum_{(i,j) \in E} [\mat{P}]_{ij} f_i f_j \right),
\end{align}
for some precision matrix $\mat{P}$ and $(V, E)$ is the graph induced by the sparsity pattern of $\mat{P}$, described earlier.
This naturally gives us factors of the form
\begin{align}
    \phi_i(f_i) := \exp\left(-\frac12 [\mat{P}]_{ii} f_i^2\right), \quad \text{and} \quad \phi_{ij}(f_i, f_j) := \exp\left(-[\mat{P}]_{ij} f_i f_j\right),
\end{align}
which we notice are also Gaussian up to a normalisation constant.
Since the factors are all proportional to Gaussian densities, and the family of Gaussians are closed under products and integrals, the messages and states computed by \eqref{eq:variable-to-factor-message}--\eqref{eq:state-update} are also expected to be Gaussians. For simplicity, let us parameterise the variable-to-factor messages \eqref{eq:variable-to-factor-message} as
\begin{align}\label{eq:var-to-fac-message}
    m_{f_i \rightarrow \phi_{ij}}(f_i) = \exp\left(-\frac12 \alpha_{ij} f_i^2 - \beta_{ij} f_i \right),
\end{align}
and the factor-to-variable messages \eqref{eq:factor-to-variable-message} as
\begin{align}\label{eq:fac-to-var-message}
    m_{\phi_{ij} \rightarrow f_i}(f_i) = \exp\left(-\frac12 a_{ij} f_i^2 - b_{ij} f_i \right).
\end{align}
We then compute the update rules for the two types of messages as follows.

\paragraph{Variable-to-factor message}
From \eqref{eq:variable-to-factor-message}, we get
\begin{align}
    m_{f_i \rightarrow \phi_{ij}}(f_i) &= \phi_i(f_i) \prod_{\substack{k \sim i \\ k \neq j}} m_{\phi_{ik} \rightarrow f_i}(f_i) \propto \exp \left(-\frac12 \Bigg(P_{ii} + \sum_{\substack{k \sim i \\ k \neq j}} a_{ik}\Bigg) f_i^2 - \sum_{\substack{k \sim i \\ k \neq j}} b_{ik} f_i\right),
\end{align}
giving us the following updates on the parameters $\alpha_{ij}$ and $\beta_{ij}$
\begin{align}\label{eq:var-to-fac-update}
    \alpha_{ij} \leftarrow P_{ii} + \sum_{\substack{k \sim i \\ k \neq j}} a_{ik}, \qquad \beta_{ij} \leftarrow \sum_{\substack{k \sim i \\ k \neq j}} b_{ik}.
\end{align}

\paragraph{Factor-to-variable message}
Now using \eqref{eq:factor-to-variable-message}, we have
\begin{align}
    m_{\phi_{ij} \rightarrow f_i}(f_i) &= \int \phi_{ij}(f_i, f_j) \mu_{f_j \rightarrow \phi_{ij}}(f_j) \mathrm{d}f_j \\
    &\propto \int \exp\Bigg(-\frac12 \alpha_{ij} f_j^2 - \beta_{ij} f_j - P_{ij} f_i f_j\Bigg) \mathrm{d}f_j \\
    &\propto \int \exp\Bigg(-\frac12 \alpha_{ij} \left(f_j - \frac{\beta_{ij}}{\alpha_{ij}}\right)^2 - P_{ij} f_i f_j\Bigg) \mathrm{d}f_j \\
    &= \int \exp\Bigg(-\frac12 \alpha_{ij} \left(f_j - \frac{\beta_{ij}}{\alpha_{ij}}\right)^2 - P_{ij} f_i \left(f_j - \frac{\beta_{ij}}{\alpha_{ij}}\right) - \frac{\beta_{ij}P_{ij}}{\alpha_{ij}} f_i \Bigg) \mathrm{d}f_j \\
    &\propto e^{- \frac{\beta_{ij}P_{ij}}{\alpha_{ij}} f_i} \int \exp\Bigg(-\frac12
    \begin{pmatrix}
        f_i \\
        f_j - \frac{\beta_{ij}}{\alpha_{ij}}
    \end{pmatrix}^\top
    \begin{pmatrix}
        0 & P_{ij} \\
        P_{ij} & \alpha_{ij}
    \end{pmatrix}
    \begin{pmatrix}
        f_i \\
        f_j - \frac{\beta_{ij}}{\alpha_{ij}}
    \end{pmatrix}
    \Bigg) \mathrm{d}f_j \\
    &=  e^{- \frac{\beta_{ij}P_{ij}}{\alpha_{ij}} f_i} \int \exp\Bigg(-\frac12
    \begin{pmatrix}
        f_i \\
        f_j - \frac{\beta_{ij}}{\alpha_{ij}}
    \end{pmatrix}^\top
    \begin{pmatrix}
        -\alpha_{ij}P_{ij}^{-2} & P_{ij}^{-1} \\
        P_{ij}^{-1} & 0
    \end{pmatrix}^{-1}
    \begin{pmatrix}
        f_i \\
        f_j - \frac{\beta_{ij}}{\alpha_{ij}}
    \end{pmatrix}
    \Bigg) \mathrm{d}f_j \\
    &\propto \exp \left(- \frac{\beta_{ij}P_{ij}}{\alpha_{ij}} f_i + \frac12 \frac{P_{ij}^2}{\alpha_{ij}} f_i^2 \right).
\end{align}
where we used the Gaussian marginalisation rule to yield the last line.
Thus, we have the following update rule on the parameters $a_{ij}$ and $b_{ij}$ determining the factor-to-variable messages
\begin{align}\label{eq:fac-to-var-update}
    a_{ij} \leftarrow -\frac{P_{ij}^2}{\alpha_{ij}}, \qquad b_{ij} \leftarrow \frac{\beta_{ij}P_{ij}}{\alpha_{ij}}.
\end{align}

In the case of Gaussian models with non-zero means, we have the following slight modification of the model
\begin{align}
    p(\vec{f}) &\propto \exp\left(-\sum_{i \in V} \left(\frac12 P_{ii} f_i^2 - h_i f_i\right) - \sum_{(i,j) \in E} P_{ij} f_i f_j \right),
\end{align}
for some vector $\vec{h} \in \mathbb{R}^N$, giving us factors of the form
\begin{align}
    \phi_i(f_i) := \exp\left(-\frac12 P_{ii} f_i^2 + h_i f_i\right), \quad \text{and} \quad \phi_{ij}(f_i, f_j) := \exp\left(-P_{ij} f_i f_j\right).
\end{align}
This only changes the variable-to-factor message update rule \eqref{eq:var-to-fac-update} to
\begin{align}
    \alpha_{ij} \leftarrow P_{ii} + \sum_{\substack{k \sim i \\ k \neq j}} a_{ik}, \qquad \beta_{ij} \leftarrow -h_i + \sum_{\substack{k \sim i \\ k \neq j}} b_{ik}.
\end{align}
The factor-to-variable message update rules \eqref{eq:fac-to-var-update} remain unchanged.

\subsection{Re-weighted message updates}
Using the re-weighted scheme \eqref{eq:reweighted-variable-to-factor-message}--\eqref{eq:reweighted-factor-to-variable-message} of \cite{wiegerinck2002fractional}, if we now parameterise the {\em fractional messages} by
\begin{align}
    m_{f_i \rightarrow \phi_{ij}}(f_i)^{\frac1c} = \exp\left(-\frac12 \alpha_{ij} f_i^2 - \beta_{ij} f_i \right), \quad m_{\phi_{ij} \rightarrow f_i}(f_i)^{\frac1c} = \exp\left(-\frac12 a_{ij} f_i^2 - b_{ij} f_i \right),
\end{align}
then the update rules on its coefficients change as
\begin{align}
    \alpha_{ij} &\leftarrow P_{ii} + c\sum_{\substack{k \sim i \\ k \neq j}} a_{ik} + (c-1)a_{ij}, \\
    \beta_{ij} &\leftarrow -h_i + c\sum_{\substack{k \sim i \\ k \neq j}} b_{ik} + (c-1) b_{ij}, \\
    a_{ij} &\leftarrow -\frac{(P_{ij}/c)^2}{\alpha_{ij}}, \label{eq:a_update} \\
    b_{ij} &\leftarrow \frac{\beta_{ij}(P_{ij}/c)}{\alpha_{ij}}. \label{eq:b_update}
\end{align}
This can be checked by direct computation. Intuitively, for $c \in \mathbb{N}$, the update rule \eqref{eq:reweighted-variable-to-factor-message}--\eqref{eq:reweighted-factor-to-variable-message} can be understood as updating a fraction of the message each time, where each update is allowed to use information from the remaining fraction of its own message. From a variational inference perspective, this can also be understood in terms of minimising the $\alpha$-divergence between the variational and true distributions with $\alpha = 2/c - 1$ \parencite{wiegerinck2002fractional, minka2004power}.

The marginal distribution at $f_i$ is then
$p(f_i) = \mathcal N(f_i | \mu_i, \sigma_i^2)$, where
\begin{align}
    &\mu_i = s_i / \sigma_i^{-2}, \label{eq:reweighted_extract_marginals-mean} \\
    &s_i = h_i + c \sum_{k \sim i} b_{ki} , \quad \text{and } \\
    &\sigma_i^{-2} = P_{ii} + c \sum_{k \sim i} a_{ki} \label{eq:reweighted_extract_marginals-precision}.
\end{align}

\section{Complete Description of Our Message-Passing Algorithm} \label{app:full_algorithm}
\Cref{alg:our_message_passing} gives full details of the message-passing algorithm we use in this work.
This is an extension of the algorithm introduced by \textcite{Ruozzi2013} (in turn a generalisation of \textcite{wiegerinck2002fractional}), with the addition of multigrid, damping, and early stopping

\begin{algorithm}[H]
    \caption{Re-weighted message passing with multigrid, damping, and early stopping}
    \label{alg:our_message_passing}
    \begin{algorithmic}
        \Procedure{Multigrid MP}{levels}
            \For{level in levels}
                \If{first level}
                    \State $m_{ij} = (0, 10^{-8})$ for $i,j \in $ level
                \Else
                    \State $m_{ij} = \Call{Upscale Messages}{m_{ij}, \text{previous level}, \text{level}}$
                \EndIf
                \State $m_{ij} =$ \Call{Iterate MP Until Convergence}{level, $m_{ij}$}
            \EndFor
            \State \Return $p(f_i) = \Call{Compute Marginal}{
                                c, \{m_{ki} : f_k \sim f_i \}
                            }$
                    for all $i \in $ last level
        \EndProcedure
        \State
        \Procedure{Iterate MP Until Convergence}{level, $m^{t=0}_{ij}$}
            \State $\{f_i\}_{i=1}^p$, $\{\phi_i\}_{i=1}^p$, $\{\phi_{ij}\}_{i,j=1}^p =$ create factor graph at level
            \Comment $p =$ number of grid points at level
            \For{$t \in \{1, \ldots T\}$}
                \For{$f_i$ in the graph}
                    \For{$f_j \in \{f_j \sim f_i \}$}
                        \State $m' = \Call{Compute Outgoing Message}{
                            c,
                            \phi_i,
                            \{ (\phi_{ki}, m^{t-1}_{ki}) : \{f_k \sim f_i \} \backslash f_j \}
                        }$
                        \State $m^{t+1}_{ij} = (1 - \eta) m^{t}_{ij} + \eta m'$
                    \EndFor
                \EndFor

                \OLIf{\Call{Early Stop}{$\{m^{1}_{ij}, m^{2}_{ij}, m^{t-1}_{ij}, m^{t}_{ij}\}_{\text{all } i,j}$}}{\textbf{break}}
            \EndFor
            \State \Return $\{m_{ij}^T\}_{\text{all }i,j}$
        \EndProcedure
    \end{algorithmic}
\end{algorithm}

\begin{figure}[H]
    \centering
    \includegraphics{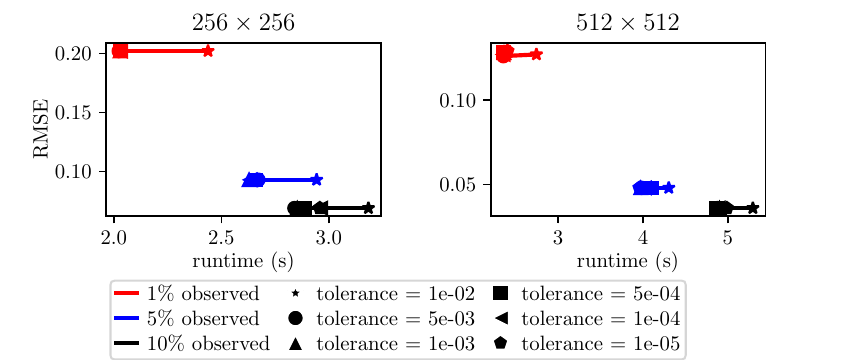}
    \caption[Convergence of message passing when varying $c$.]{
        Effect of the early stopping hyperparameter on the error and runtime of 3D-Var.
        In particular, we search over the \emph{tol} hyperparameter of the JAXopt L-BFGS optimiser.
        We consider two grid sizes, $256 \times 256$ and $512 \times 512$, and $1$\%, $5$\%, and $10$\% of the grid being observed.
        Based on these results we select \emph{tol} $=1 \times 10^{-3}$.
    }
    \label{fig:lbfgs_early_stopping_grid}
\end{figure}

The functions that the algorithm depends on are defined as follows:
\begin{itemize}
    \item $\Call{Compute Outgoing Message}{c, \phi_i, \{ (\phi_{ki}, m^{t-1}_{ki}) : \{f_k \sim f_i \} \backslash f_j \}}
     = (a_{ij}', b_{ij}')$
    where $a_{ij}'$ and $b_{ij}'$ are defined by \cref{eq:a_update,eq:b_update} respectively in \cref{app:message_passing_background}.

    \item $\Call{Compute Marginal}{c, \{m_{ki} : f_k \sim f_i \}}$ is given by \cref{eq:reweighted_extract_marginals-mean} -- \eqref{eq:reweighted_extract_marginals-precision}.

    \item We stop iterating if the following inequality holds:
    \begin{equation*}
        \Call{Early Stop}{\{m^{1}_{ij}, m^{2}_{ij}, m^{t-1}_{ij}, m^{t}_{ij}\}_{\text{all } i,j}} =
        \operatorname{mean}(\operatorname{abs}(m^{t}_{ij} - m^{t-1}_{ij}))
        < \tau \operatorname{mean}(\operatorname{abs}(m^{1}_{ij} - m^{2}_{ij})),
    \end{equation*}
    where the mean is taken over all $i,j$, and $\tau = 0.001$ based on \cref{fig:lbfgs_early_stopping_grid}.
    \item \Call{Upscale Messages}{}: The edge values on the finer grid are initialised to the edge values of the preceding coarser grid.
        This is equivalent to initialising the marginals of the finer grid to those of the coarser grid, but avoids explicitly calculating the marginals.
\end{itemize}

\section{Details on the Finite Difference Discretisation}\label{app:discretisation}
In this section, we provide details on the discretisation of the SPDE \eqref{eq:matern-spde}, from which we derive our graphical model to which message passing is then applied.
We only consider finite difference discretisation. However, it would also be possible to use finite elements.

\subsection{Discretisation of the Differential Operator}
First, consider the operator $\mathcal{L} := (\kappa^2 - \Delta)^{\alpha/2}$ defining the LHS of \eqref{eq:matern-spde}. This is a differential operator provided $\alpha/2 \in \mathbb{N}$. For simplicity, consider the system in one spatial dimension and $\alpha/2 = 1$. Taking a 1D grid with equal spacing $\Delta x = h$, a finite difference discretisation of the LHS of \eqref{eq:matern-spde} reads
\begin{align}\label{eq:discretised-operator}
    (\kappa^2 - \Delta) f(x_i) \approx \kappa^2  f_i - \frac{f_{i+1} + f_{i-1} - 2 f_{i}}{h^2},
\end{align}
where $\{x_i\}_{i=0}^N$ are the grid points and we used the shorthand notation $f_{i} := f(x_i)$. This gives us a finite dimensional approximation $(\kappa^2 - \Delta) f \approx \mat{L} \vec{f}$, where $\vec{f} = (f_0, \ldots, f_N)^\top$ and the matrix $\mat{L}$ has entries
\begin{align}
    [\mat{L}]_{ij} =
    \begin{cases}
        \kappa^2 + 2/h^2, &\quad \text{if} \quad i = j, \\
        -1/h^2, &\quad \text{if} \quad |i-j| = 1, \\
        0, &\quad \text{otherwise}
    \end{cases}
\end{align}
for $i, j = 1, \ldots, N-1$, which is sparse and banded, with bandwidth equal to one. On the boundaries of the domain, $i,j \in \{0, N\}$, we impose the Dirichlet condition, which sets $f_0 = f_N = 0$, and therefore we can exclude them from the vector $\vec{f}$, resulting in a $(N-2)$-dimensional linear system. We may also consider the periodic boundary condition, which sets $f_{-1} = f_N$ and $f_{N+1} = f_0$, when such terms (called ghost points) appear in equation \eqref{eq:discretised-operator}.

On a 2D $N_x \times N_y$ grid, one can apply a similar discretisation to get the approximation $\mathcal{L} f := (\kappa^2 - \Delta) f \approx \mat{L} \vec{f}$, where now, $\vec{f}$ is a vector in $\mathbb{R}^{N_xN_y}$ and $\mat{L} \in \mathbb{R}^{N_xN_y \times N_xN_y}$. We can also obtain higher-order powers of the operator $\mathcal{L}$ by taking $\mathcal{L}^n f \approx \mat{L}^n \vec{f}$. In our experiments, we use the \texttt{findiff} package \parencite{findiff} to perform the discretisation of the operator $\mathcal{L}$.

\subsection{White Noise Discretisation} \label{app:white-noise-discretisation}
For the Gaussian white-noise term $\mathcal{W}$ in \eqref{eq:matern-spde} (assuming that we are in a compact domain $C \subset \mathbb{R}^2$), we claim that this can be approximated by a stochastic process $\mathcal{W}^N : C \rightarrow \mathbb{R}$ of the form
\begin{align}
    \mathcal{W}^N(x) = \sum_{i=1}^N \frac{z_i}{\sqrt{\Delta x \Delta y}} \,\mathbf{1}_{C_i}(x),
\end{align}
where
\begin{align}
    \boldsymbol{z}^N = (z_1, \ldots, z_N) \sim \mathcal{N}(0, I_N)
\end{align}
and $\{C_i\}_{i=1}^N$ denotes a finite-difference discretised rectangular cell in $C$ whose volume $\Delta x \Delta y$ vanishes as $N \rightarrow \infty$.
First, we formally define the spatial white-noise process as follows.
\begin{definition}[\cite{lototsky2017stochastic}]
    Given a probability triple $(\Omega, \mathcal{F}, \mathbb{P})$, a random element $\mathcal{W} : \Omega \rightarrow L^2(C, \mathbb{R})^*$ (the space of continuous functionals on $L^2(C, \mathbb{R})$) is called a (zero-mean) Gaussian white-noise process in $L^2(C, \mathbb{R})$ if it satisfies the following properties:
    \begin{enumerate}
        \item For every $f \in L^2(C, \mathbb{R})$, we have $\mathcal{W}f = 0$.
        \item For every $f, g \in L^2(C, \mathbb{R})$, we have $\mathbb{E}[\mathcal{W}f, \mathcal{W}g] = \left<f, g\right>_{L^2}$.
    \end{enumerate}
\end{definition}

To justify that the process $\mathcal{W}^N$ approximates $\mathcal{W}$, for every $f \in L^2(C, \mathbb{R})$, we have
\begin{align}
    \left<\mathcal{W}^N, f\right>_{L^2} = \sum_{i=1}^N \frac{z_i}{\sqrt{\Delta x \Delta y}} \int \int_{C_i} f(x) \mathrm{d}x \mathrm{d}y.
\end{align}
Now for small $|C_i| = \Delta x \Delta y$, we have the approximation
\begin{align}
    \int \int_{C_i} f(x) \mathrm{d}x \mathrm{d}y \approx f(x_i) \Delta x \Delta y,
\end{align}
for some arbitrary $x_i \in C_i$ (e.g. the central point of $C_i$). Thus, we have the approximation
\begin{align}
    \left<\mathcal{W}^N, f\right>_{L^2} &\approx \sum_{i=1}^N \frac{z_i f(x_i)}{\sqrt{\Delta x \Delta y}} \Delta x \Delta y = \sum_{i=1}^N z_i f(x_i) \sqrt{\Delta x \Delta y}
\end{align}
and we see that the random variable $\left<\mathcal{W}^N, h\right>_{L^2}$ is Gaussian with moments
\begin{align}
    \mathbb{E}\left[\left<\mathcal{W}^N, f\right>_{L^2}\right] &= \sum_{i=1}^N \underbrace{\mathbb{E}[z_i]}_{=0} f(x_i) \sqrt{\Delta x \Delta y} = 0 \\
    \mathbb{E}\left[\left<\mathcal{W}^N, f\right>_{L^2} \left<\mathcal{W}^N, g\right>_{L^2}\right] &= \sum_{i=1}^N \sum_{j=1}^N \underbrace{\mathbb{E}[z_i z_j]}_{= \delta_{ij}} f(x_i) g(x_j) \Delta x \Delta y = \sum_{i=1}^N f(x_i) g(x_i) \Delta x \Delta y.
\end{align}
Taking $N \rightarrow \infty$, these converge as
\begin{align}
    &\mathbb{E}\left[\left<\mathcal{W}^N, f\right>_{L^2}\right] \rightarrow 0 \\
    &\mathbb{E}\left[\left<\mathcal{W}^N, f\right>_{L^2} \left<\mathcal{W}^N, g\right>_{L^2}\right] = \sum_{i=1}^N f(x_i) g(x_i) \Delta x \Delta y \rightarrow \int_{C} f(x) g(x) \mathrm{d} x \mathrm{d} y = \left<f, g\right>_{L^2}.
\end{align}
Note that the latter convergence follows from the definition of Riemann integration.
Thus, the moments of $\left<\mathcal{W}^N, f\right>_{L^2}$ converge to the moments of $\mathcal{W}f$ as $N \rightarrow \infty$ and since $f, g \in L^2(C, \mathbb{R})$ were chosen arbitrarily, we have the convergence in law
\begin{align}
    \mathcal{W}^N \rightarrow \mathcal{W}.
\end{align}

Putting this together, we find a discretised representation of \eqref{eq:matern-spde} in the form
\begin{align}
    \mat{L} \vec{f} = \frac{1}{\sqrt{\Delta x \Delta y}} \vec{z}, \quad \vec{z} \sim \mathcal{N}(0, I).
\end{align}
In practice, we wish to have control over the marginal variances of the process $\vec{f}$ that can be tuned on the data. To achieve this, we modify the expression slightly as
\begin{align}
    \mat{L} \vec{f} = \sqrt{\frac{\sigma^2 q}{\Delta x \Delta y}} \vec{z}, \quad \vec{z} \sim \mathcal{N}(0, I).
\end{align}
where $q$ is a constant that takes the form
\begin{align}
    q := \frac{(4\pi)^{d/2}\kappa^{2\nu}\Gamma(\nu+d/2)}{\Gamma(\nu)}, \label{eq:q}
\end{align}
This allows $\mathbb{E}[f_i^2] = \sigma^2$ for all $i = 1, \ldots, N$ and we can subsequently treat $\sigma$ as a tunable parameter. We note that expression \eqref{eq:q} can be obtained from the limit $q^{-1} = \lim_{\vec{x}' \rightarrow \vec{x}} k(\vec{x}, \vec{x}') / \sigma^2$, where $k(\vec{x}, \vec{x}')$ is the Matérn kernel
\begin{equation}
    k(\vec{x}, \vec{x}') = \sigma^2 \frac{2^{1-\nu}}{\Gamma(\nu)} \left(\sqrt{2\nu} \frac{\|\vec{x}-\vec{x}'\|}{\ell}\right)^\nu K_\nu\left(\sqrt{2\nu} \frac{\|\vec{x}-\vec{x}'\|}{\ell}\right).
\end{equation}
Here, $K_\nu$ is the modified Bessel function of the second kind, $\Gamma$ is the Gamma function, and the hyperparameters $\nu, \sigma^2$ and $\ell$ govern the smoothness, variance and lengthscale of the corresponding process respectively.

\subsection{Extension to Spatiotemporal Systems} \label{app:spatiotemporal_extension}
Beyond the spatial setting considered here, we can also construct GMRF representations of linear spatiotemporal SPDEs, such as the 1D stochastic heat equation
\begin{align}\label{eq:stochastic-heat-eqn}
    \frac{\partial u}{\partial t} = \nu \Delta u + \sigma \mathcal{W},
\end{align}
for some coefficients $\nu, \sigma > 0$.
In general, we can describe a first-order-in-time discretisation of a temporally evolving system in the form
\begin{align}\label{eq:discretised-dynamics}
\mathbf{A}\vec{u}^{n+1} = \mat{B}\vec{u}^{n} + \frac{\sigma}{\sqrt{\Delta t\Delta x}} \,\vec{z}^{n+1}, \quad \vec{z}_{n+1} \sim \mathcal{N}(0, I), \quad n = 0, \ldots, N-1,
\end{align}
for some time step $n$, where $\mat{A}$ and $\mat{B}$ are some matrices determined by the numerical method used for time-discretisation, such as the Crank-Nicolson scheme. Let us also assume a random initial condition distributed according to a Gaussian
\begin{align}
    \vec{u}^0 \sim \mathcal{N}(\vec{m}, \mat{P}^{-1}),
\end{align}
for some mean $\vec{m}$ and precision $\mat{P}$. Assuming that we can write $\mat{P} = (\Delta t\Delta x / \sigma^2) \,\mat{L}^\top\mat{L}$ for some invertible matrix $\mat{L}$, the initial condition can be re-expressed as
\begin{align}\label{eq:init-cond}
    \mat{L} \vec{u}_0 = \mat{L} \vec{m} + \frac{\sigma}{\sqrt{\Delta t\Delta x}}\vec{z}_0, \quad \vec{z}_0 \sim \mathcal{N}(0, \mat{I}).
\end{align}
Then, combining equations \eqref{eq:discretised-dynamics} and \eqref{eq:init-cond}, we get a large matrix-vector system of the form
\begin{align}
\begin{pmatrix}
\mat{L} & 0 & 0 & \cdots & 0 & 0 \\-\mat{B} & \mat{A} & 0 & \cdots & 0 & 0 \\0 & -\mat{B} & \mat{A} & \cdots & 0 & 0\\& \vdots && \ddots & \vdots & \vdots \\0 & 0 & 0 & \cdots & -\mat{B} & \mat{A}
\end{pmatrix}
\begin{bmatrix}
\boldsymbol{u}^0 \\ \boldsymbol{u}^1 \\ \boldsymbol{u}^2 \\ \vdots \\ \boldsymbol{u}^N
\end{bmatrix}
=
\begin{bmatrix}
\mat{L}\boldsymbol{m} \\ 0 \\ 0 \\ \vdots \\ 0\end{bmatrix}
+ \frac{\sigma}{\sqrt{\Delta t\Delta x}}
\begin{bmatrix}
\boldsymbol{z}^0 \\ \boldsymbol{z}^1 \\ \boldsymbol{z}^2 \\ \vdots \\ \boldsymbol{z}^N
\end{bmatrix}.
\end{align}
Now, denoting the matrix on the LHS by $\mat{M}$, we have the solution
\begin{align}
\begin{bmatrix}
\boldsymbol{u}^0 \\ \boldsymbol{u}^1 \\ \boldsymbol{u}^2 \\ \vdots \\ \boldsymbol{u}^N
\end{bmatrix}
\sim
\mathcal{N}\left(\mat{M}^{-1}\begin{bmatrix}
\mat{L}\boldsymbol{m} \\ 0 \\ 0 \\ \vdots \\ 0\end{bmatrix}, \, \frac{\sigma^2}{\Delta t \Delta x}(\mat{M}^\top\mat{M})^{-1}\right),
\end{align}
which is a GMRF if the precision $\mat{M}^\top\mat{M}$ is sparse. In natural parameterisation, this has a neater expression, with precision matrix
\begin{align}
    \mat{Q} = \frac{\Delta t \Delta x}{\sigma^2} \mat{M}^\top \mat{M}
\end{align}
and shift vector
\begin{align}
    \boldsymbol{b} := \mat{P}\mat{M}^{-1}
    \begin{bmatrix}
    \mat{L}\boldsymbol{m} \\ 0 \\ 0 \\ \vdots \\ 0
    \end{bmatrix}
    = \frac{\Delta t \Delta x}{\sigma^2}\mat{M}^{\top}
    \begin{bmatrix}
    \mat{L}\boldsymbol{m} \\ 0 \\ 0 \\ \vdots \\ 0
    \end{bmatrix}
    =  \frac{\Delta t \Delta x}{\sigma^2}
    \begin{bmatrix}
    \mat{L}^\top\mat{L}\boldsymbol{m} \\ 0 \\ 0 \\ \vdots \\ 0
    \end{bmatrix}
    =
    \begin{bmatrix}
    \mat{P}\vec{m} \\ 0 \\ 0 \\ \vdots \\ 0
    \end{bmatrix}.
\end{align}

\begin{example}
    Using the Crank-Nicolson scheme, one can discretise the system \eqref{eq:stochastic-heat-eqn} as
    \begin{align}
        &\quad \boldsymbol{u}^{n+1} = \boldsymbol{u}^{n} + \frac{\Delta t}{2} \left(\boldsymbol{L}\boldsymbol{u}^{n+1} + \boldsymbol{L}\boldsymbol{u}^{n}\right) + \sqrt{\frac{\sigma^2 \Delta t}{\Delta x}} \boldsymbol{z}^{n+1} \\ &\Rightarrow \mathbf{A}\vec{u}^{n+1} = \mat{B}\vec{u}^{n} + \frac{\sigma}{\sqrt{\Delta t\Delta x}}\vec{z}^{n+1},
    \end{align}
    where $\vec{L}$ is the discretised Laplacian operator and
    \begin{align}
        \mat{A} := \frac{1}{\Delta t}\mat{I} - \frac12 \boldsymbol{L}, \qquad \mat{B} := \frac{1}{\Delta t}\mat{I} + \frac12 \boldsymbol{L}.
    \end{align}
\end{example}

\newpage
\section{Additional Results} \label{app:additional_results}
\subsection{Choice of Hyperparameters} \label{app:grid_search}
\begin{figure}[H]
    \centering
    \includegraphics{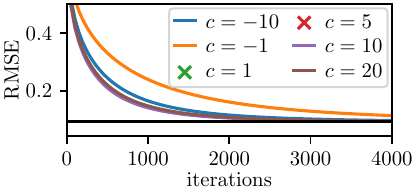}
    \caption[Convergence of message passing when varying $c$.]{
        Convergence of message passing for different values of $c$ for a grid size of $256 \times 256$.
        For each $c$ we plot the result for the best $\eta$.
        $\bm{\times}$ indicates that the algorithm diverged.
        The black horizontal line shows INLA.
    }
    \label{fig:mp_hyperparameters}
\end{figure}

We perform a grid search to set the $c$ and learning rate hyperparameters, which are crucial to the convergence of the algorithm.
If $c$ has too small a magnitude, or if the learning rate is too high, the algorithm will diverge.
We perform the grid search on simulated data of various grid sizes.
\cref{fig:mp_hyperparameters} highlights the speed of convergence of several values of $c$, and \cref{tab:full_grid_search} gives the full results.
Based on these results we choose $c=10$ and $\eta=0.6$ for the rest of our experiments.

\begin{table}[H]
    \caption{
        Grid search over $c$ and $\eta$ for various grid sizes and an observation density of 5\%.
        We report the ratio $\frac{\text{RMSE of message passing}}{\text{RMSE of INLA}}$ after $4000$ iterations, with lower values being better.
        ``-'' indicates that the method diverged for that combination of hyperparameters.
    }
    \label{tab:full_grid_search}
    \centering
    \begin{tabular}{llccccccc}
        \toprule
        & & \multicolumn{7}{c}{$\bm{c}$} \\
        \cline{3-9}
        \textbf{grid size} & $\bm{\eta}$ & -10 & -2 & -1 & 1 & 5 & 10 & 20 \\
        \midrule
        \multirow[t]{3}{*}{$128 \times 128$} & 0.6 & 3.83 & 3.88 & 3.99 & - & - & \textbf{3.82} & \textbf{3.82} \\
        & 0.7 & - & 3.85 & 3.92 & - & - & - & - \\
        & 0.8 & - & - & - & - & - & - & - \\
       \cline{1-9}
       \multirow[t]{3}{*}{$256 \times 256$} & 0.6 & 2.04 & 2.19 & 2.40 & - & - & \textbf{1.98} & 2.00 \\
        & 0.7 & - & 2.11 & 2.28 & - & - & - & - \\
        & 0.8 & - & - & - & - & - & - & - \\
       \cline{1-9}
       \multirow[t]{3}{*}{$512 \times 512$} & 0.6 & 1.30 & 1.68 & 2.18 & - & - & \textbf{1.13} & 1.17 \\
        & 0.7 & - & 1.49 & 1.90 & - & - & - & - \\
        & 0.8 & - & - & - & - & - & - & - \\
        \bottomrule
    \end{tabular}
\end{table}
\begin{figure}[H]
    \centering
    \includegraphics{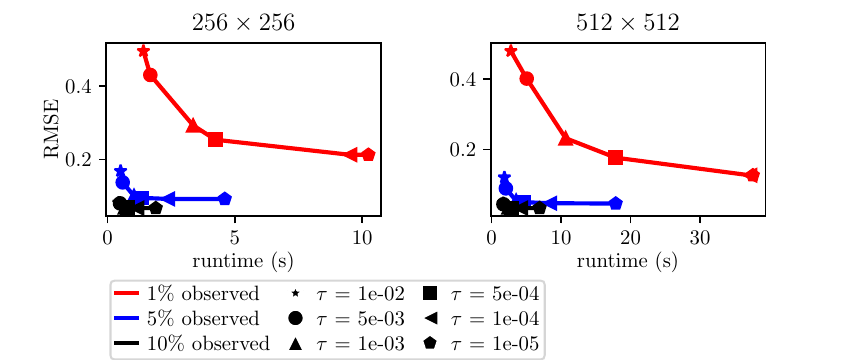}
    \caption{
        Effect of the early stopping hyperparameter, $\tau$, on the error and runtime of multigrid message passing.
        We consider two grid sizes, $256 \times 256$ and $512 \times 512$, and $1$\%, $5$\%, and $10$\% of the grid being observed.
        Based on these results we select $\tau = 1 \times 10^{-3}$.
    }
    \label{fig:mp_early_stopping_grid}
\end{figure}

\subsection{Temperature Data}
\begin{figure}[H]
    \centering
    \includegraphics[width=0.5\textwidth]{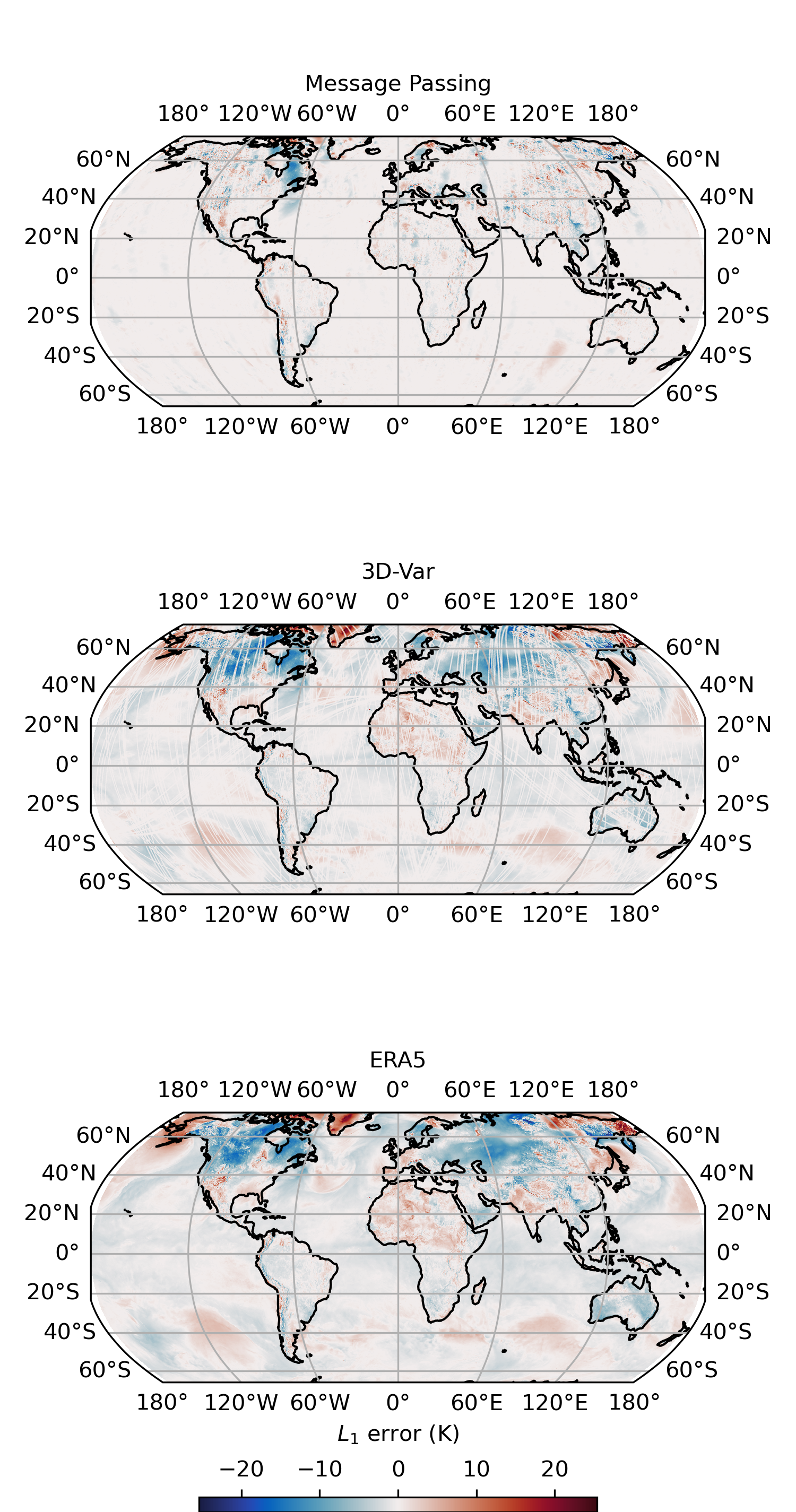}
    \caption{
        $L_{1}$ errors for the message passing, 3D-Var and the prior (ERA5) against the high-resolution Met Office Unified Model temperature data.
    }
    \label{fig:temperature_data_l1}
\end{figure}

\cref{fig:temperature_data_l1} showcases the $L_{1}$ error for our message passing approach, our 3D-Var implementation and for the prior mean field. The mean of ERA5 surface temperature from 06:00UTC $1^{st}$ January for the years 2000 to 2019 forms the prior mean field. This prior mean field is mapped to the high-resolution grid and the high-resolution data is valid for 06:00 UTC $1^{st}$ January 2020. For the DA approaches we see an improvement when compared to the error of the prior mean, with the message-passing approach performing better. The message passing implementation benefits from the multigrid approach with observations from the satellite tracks being propagated away from the locations. It should be noted that a hyperparameter tuning of the 3D-Var implementation, which is not carried out here, could improve its performance.

\section{Experiment Details} \label{app:experiment_details}
All experiments are performed on a 24-core AMD Threadripper 3960X CPU and an Nvidia RTX 3090 GPU.
We use JAX version 0.4.23, jaxlib 0.4.23+cuda12.cudnn89.

\paragraph{Message passing hyperparameters}
Unless otherwise mentioned we use $c=10$ and $\sigma = 0.6$, as selected in the grid search in \cref{app:grid_search}.
We set $T=10,000$, though this many iterations is rarely used because of the early stopping.

\paragraph{3D-Var hyperparameters}
We use the default hyperparameters of L-BFGS as implemented in JAXopt version 0.8.3.
In particular max iterations $=500$, stopping tolerance $=10^{-3}$, and zoom line search.

\paragraph{R-INLA}
We use version 23.9.9.

\paragraph{Synthetic data}
We generate the data by simulating the SPDE given in \cref{eq:matern-spde}, using a finite difference solver.
We set $\alpha=2$, $\kappa = \sqrt{2\nu} / l$, with $\nu = 1$, $l=0.15$, and $\sigma=1.1$.
We then use this SPDE as the prior for inference.

\paragraph{Global temperature data}
Message passing uses three multigrid levels of grid sizes $625 \times 375$, $1250 \times 750$ and $2500 \times 1500$. \Cref{tab:satellites} lists the satellites used to generate the observation locations.
To specify the prior we set $\alpha=2$, $\kappa = \sqrt{2\nu} / l$, with $\nu = 1$, $l=0.2$, and $\sigma=1.9$.

\begin{table}[H]
    \caption{
        List of satellites used to generate the observation locations in the global temperature data example.
    }
    \label{tab:satellites}
    \centering
    \small
    \begin{tabular}{cccc}
        \toprule
        NOAA 15 & DMSP 5D-3 F16 (USA 172) &  NOAA 18 & METEOSAT-9 (MSG-2) \\
        EWS-G1 (GOES 13) & DMSP 5D-3 F17 (USA 191) & FENGYUN 3A & FENGYUN 2E  \\
        NOAA 19  & GOES 14 & DMSP 5D-3 F18 (USA 210) & EWS-G2 (GOES 15) \\
        COMS 1  & FENGYUN 3B & SUOMI NPP & FENGYUN 2F \\
        METEOSAT-10 (MSG-3) & METOP-B & FENGYUN 3C & METEOR-M 2 \\
        HIMAWARI-8 & FENGYUN 2G  & METEOSAT-11 (MSG-4) & ELEKTRO-L 2 \\
        HIMAWARI-9 & GOES 16 & FENGYUN 4A & CYGFM05 \\
        CYGFM04 & CYGFM02 & CYGFM01 & CYGFM08 \\
        CYGFM07 & CYGFM03 & FENGYUN 3D & NOAA 20 \\
        GOES 17 & FENGYUN 2H & METOP-C & GEO-KOMPSAT-2A \\
        METEOR-M2 2 & ARKTIKA-M 1 & FENGYUN 3E & GOES 18 \\
        NOAA 21 (JPSS-2) & METEOSAT-12 (MTG-I1) & TIANMU-1 03 & TIANMU-1 04\\
        TIANMU-1 05 & TIANMU-1 06 & METEOR-M2 3 & TIANMU-1 07 \\
        TIANMU-1 08 & TIANMU-1 09 & TIANMU-1 10 & FENGYUN 3F \\
        TIANMU-1 11 & TIANMU-1 20 & TIANMU-1 21 & TIANMU-1 22 \\
        TIANMU-1 15 & TIANMU-1 16 & TIANMU-1 17 & TIANMU-1 18 \\
        \bottomrule
    \end{tabular}
\end{table}

\end{appendix}

\end{document}